# Graph Neural Network-based Unsupervised Early Bearing Fault Detection

Xusheng Du*[1], Jiong Yu[1]

1.School of Information Science and Engineering, Xinjiang University, Urumqi 830046, China

*Abstract*—Early detection of faults is importance to avoid catastrophic accidents and ensure safe operation of machinery. A novel graph neural network-based fault detection method is proposed to build a bridge between AI and real-world running mechanical systems. First, the vibration signals, which are Euclidean structured data, are converted into graph (non-Euclidean structured data). Second, inputs the dataset together with its corresponding graph into the GNN for training, which contains graphs in each hidden layer of the network, enabling the graph neural network to learn the feature values of itself and its neighbors, and the obtained early features have stronger discriminability. Finally, determines the top-*n* objects that are difficult to reconstruct in the output layer of the GNN as fault objects. A public datasets of bearings have been used to verify the effectiveness of the proposed method. We find that the proposed method is able to accurate detect those early fault objects (mixed in the normal object area).The experiment codes are available at: https://github.com/sunrise03/Graph-Neural-Network-based-Early-Bearing-Fault-Detection

## I. Introduction

Rotating machinery is an indispensable equipment in aerospace, electric power, transportation, petroleum, chemical and other fields , which has a wide range of applications[1]. These machines consist of stator, rotor, and rolling element bearing (REB), etc. The REB is the core and fragile part of rotating machinery, which consists of inner race, ball and outer race. The operating environment of this type of equipment is harsh and complex. During long-term operation, the mechanical performance gradually decreases and easily result in failures [2]. However, the signal of early bearing fault is not obvious, and most of the existing methods are difficult to detection accurately, which can easily cause missed report. Failure to detection bearing faults in a timely manner can leading to entire system downtime, high maintenance costs and potential risk of loss of lives. Therefore, early bearing fault detection is essential to avoid catastrophic accidents and ensure the safe operation of machinery [3].

Many researchers have worked on developing early fault detection (EFD) techniques [4-9] to monitor the health of rotating machinery [10]. Among these sensor signals, scholars have conducted a lot of research on how to effectively analyze vibration signals for fault detection. This is because vibration signals usually contain rich information about the health condition of the equipment [11, 12]. In general, vibration signal-based fault detection methods can be divided into: a) time domain; b) frequency domain; c) time-frequency domain. Since degradation of rotating mechanical properties is often accompanied by changes in vibration. This causes a change in the probability distribution of the vibration signal. At the same time, the properties of the probability density distribution also changed. Therefore, time domain–based method is highly sensitive to the vibration of equipment failures and simple to understand with a clear physical meaning. The signal processing methods based on the time domain include mean, variance, mean square root, waveform, skewness, peak, margin, kurtosis, pulse and et al. [13]. The signal processing techniques based on the frequency domain are spectral analysis, higher-order spectral analysis and demodulated spectral analysis [14]. It can expose the frequency components of the signal and the energy distribution of the frequencies. It is one of the most widely used methods in vibration signal processing. The time-frequency domain signal processing techniques include short-time Fourier transform (STFT) [15], wavelet transform (WT) [16], wavelet packet decomposition (WPD) [17], empirical mode decomposition (EMD) [18] and their variants [19, 20]. These methods extract vibration signal characteristics by analyzing the generation times of different frequency signals, and the frequency components contained in different times. Ali et al. [21] used EMD energy entropy as a bearing fault feature for automatic fault classification by a neural network algorithm. Liu et al. [22] decomposed a bearing vibration signal using local mean decomposition (LMD) and calculated multi-scale entropy as a feature vector based on the decomposed signal components. Zhang et al [23] used the ensemble empirical mode decomposition (EEMD) to calculate the alignment entropy for bearing fault signals, and then used an optimized support vector machine for the detection of faults. However, these methods require complete a priori knowledge, as well as manual involvement in extracting features. This leads to low accuracy of fault detection.

With the rapid development of machine learning, deep learning has become an effective method to overcome the above-mentioned drawbacks. Deep learning-based methods automatically extract features through neural networks, resulting in features that make it easier to identify and detection fault objects. Deep learning methods can be broadly divided into supervised and unsupervised.

Supervised deep learning is mainly accomplished by constructing statistical recognition functions and combining them with typical sample training. First, by learning a large number of training samples with known class labels. Secondly, it is accomplished by simulating a discriminant function to classify samples with unknown classes [24]. Supervised fault detection methods all inevitably require labels. However, in most cases, there are lack of prior validated knowledge of the fault or untagged historical data. To solve this problem, unsupervised learning methods have been extensively studied and successfully applied in the field of fault detection [25]. Unsupervised fault detection methods utilize statistical analysis methods. Examples include constrained Boltzmann machines [26], autoencoder [27], generative adversarial network [13], etc. Classification is based on sample characteristic parameters and statistical features. This means that in the absence of any known class sample learning. Methods for analyzing sample similarity or probability density function estimates. Obtaining the intrinsic structure of the sample data. The correct classification of the samples is achieved. For example, Hu et al [28] proposed a fault detection method that combines a deep Boltzmann machine with multi-grain size forest integration. The raw features are transformed into some binary vectors. The corresponding class vectors are generated by multigranularity scanning and forest processing. Then the device health condition is obtained by forest integration. Wang et al [29] used a variation autoencoder for fault detection of industrial paper machines. Han et al. [30] proposed a deep adversarial convolutional neural network (DACNN) for rotor fault detection. However, the signal strength of early fault features is weak and disturbed by noise. The fault features extracted by the above methods are not obvious, resulting in low diagnostic accuracy.

The main contributions of this paper are:

1. This paper proposes a method for converting Euclidean structured data sets into graphs. This allows objects that were previously independent of each other to be related. We construct directed edges between objects. Using similarity values as weights for the edges. Use the objects as vertices connecting the top-*n* neighbors that are most similar to the objects.

2. In this paper, a new neural network structure is proposed by invoking graph convolutional networks. The proposed network contains graphs in each hidden layer of the network. This approach enables the neural network to learn the eigenvalues of objects and their neighbors.

3. Application of GNN to bearing failure detection. Fault detection is achieved by providing the dataset and its corresponding graphs to GNN for training purposes. To the best of our knowledge, this is the first graphical neural network model dedicated to bearing fault detection.

4. We have conducted extensive experiments on the CWRU dataset to demonstrate the effectiveness of the new method. The results show that the method can successfully detect fault objects mixed in the normal objects region.

## II. METHODOLOGY

Since the early fault characteristics of mechanical equipment are not obvious, the traditional fault detection technology is difficult to accurately identify. Aim to this problem, we specially designed a method for detect faults. The main structure of our model is the graph neural network in deep learning. The model consists of three modules: construct graph; graph neural network; fault detection.

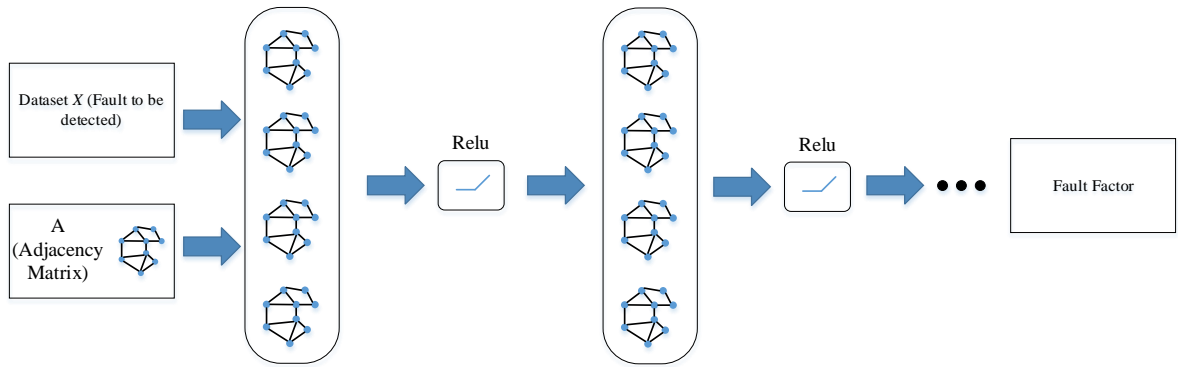

Figure.1 The entire structure of GNN for bearing fault detection

### A. Construct graph

In the Euclidean structure dataset, objects are independent of each other, and there is no connection relationship. We propose a method to convert the original unconnected relationship into a connected relationship, called construct graph.

Definition 1: Similarity

$Sim(X_i, X_j)$ is defined to measure the similarity between objects, where $X_i, X_j \in X$ and $X_i \neq X_j$. When the value of $Sim(X_i,$ $X_j$) tends to 1, the two objects are more similar. Commonly used similarity measures are Euclidean distance, cosine similarity, etc.

Definition 2: *k* neighborhood of an object $x_i$

For any object $x_i$, the *k* objects that are most similar to it are selected as its neighbors, and a directed edge pointing from $x_i$'s neighbors to $x_i$ is constructed. The set formed by $x_i$'s neighbors is called $x_i$'s neighborhood and is denoted as $N_k(x_i)$.

Definition 3: Edge weight

The weights on the edge connecting $X_j$ to $X_i$ are calculated by equation (1):

$$W(X_i, X_j) = \frac{Sim(X_i, X_j)}{\sum_{j=1}^{k} Sim(X_i, X_j)}, X_j \in N_k(X_i) \quad (1)$$

The higher the similarity between $X_j$ and $X_i$, the higher the weights on their connected edges.

Definition 4: Graph

Each object in the data set $X$ is taken as a vertex, and the vertex are combined with the connection relations between them to form the directed graph $A$.

$$Graph = (V, E, W) \quad (2)$$

We denote the generated graph by the adjacency matrix $A$. The diagonal value in $A$ is set to 1. It is worth noting that $W(X_i, X_j)$ is not necessarily equal to $W(X_j, X_i)$.

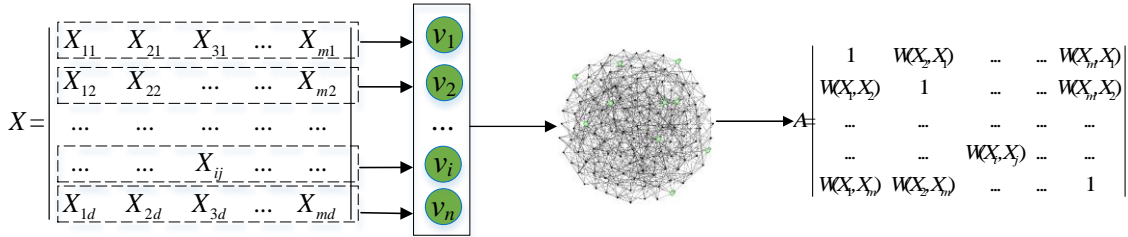

Figure. 2 Construct graph

**Algorithm1** Construct graph

**Input**: Fault to be detected $X$, the number of nearest neighbor's *k*.

**Output**: Generated graph (adjacency matrix $A$)

1. Calculate the similarity between any two objects in $X$.
3. **for** $i=1:m$ (*m* represent the number of objects in $X$)
4. **for** $j=1:m$
5. $W(X_i, X_j) = \frac{Sim(X_i, X_j)}{\sum_{j=1}^{k} Sim(X_i, X_j)}, X_j \in N_k(X_i)$
6. $A_{ji} = W(X_i, X_j)$.
7. **end**
8. **end**
10. **return** $A$.

### B. Graph neural network

Our forward model then takes the simple form:

$$Z = f(X, A) = \sigma((\sigma(XAW^{(0)})AW^{(1)}) \quad (3)$$

Here, $X$ denotes the dataset to be detected, $A$ denotes the adjacency matrix constructed from $X$, and $W$ denotes the weight matrix between hidden layers, $\sigma$ denotes the activation function. The weights are updated by batch gradient descent.

The loss function of the graph neural network is:

$$J(W,b) = \sum(L(X', Z)) = \sum \|X' - Z\|^2 \quad (4)$$

In equation (4), $X'=X*A*A$, $X$ denotes the dataset to be detected, contain normal objects and outliers, $Z$ is the output of the GNN. In this work, we perform batch gradient descent using the full dataset for every training iteration.

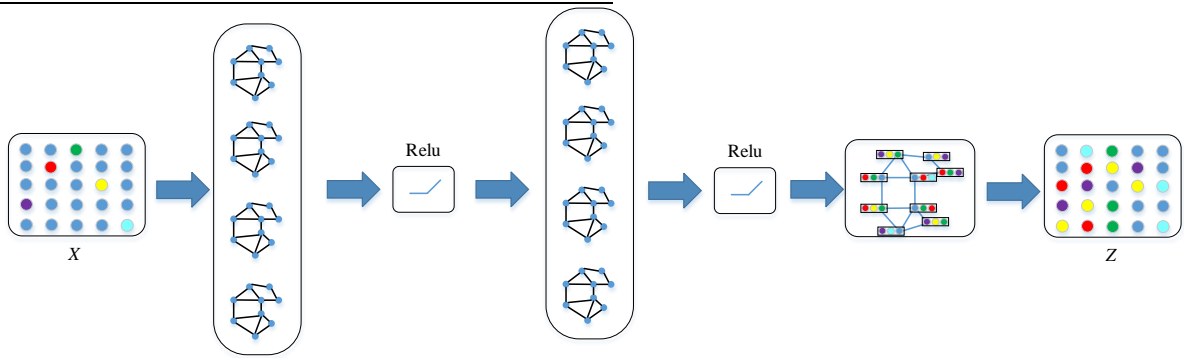

Figure. 3 Graph neural network

The GNN uses a feed-forward multilayer graph neuron network with one hidden layer sandwiched between an input layer and an output layer. GNN incorporates a graph at each layer of the network. The objects propagate their own feature value to each other based on the connection relationship between them. The data set $X$ is reconstructed by the graph neural network into a matrix $Z$, in which each object contains not only information about the object itself, but also information about the object and its neighbors.

The purpose of GNN training is to learn the feature between $X_i$ and its neighbors and reconstruct it in the output layer to minimize the mean squared error between $X'$ ($X'=X*A*A$) and $Z$. Since the majority of objects in the dataset are normal objects, normal objects will be easier to reconstruct by the GNN in order to minimize the reconstruction error of the overall dataset;

however, fault objects are difficult to reconstruct because they deviate from the majority of objects.

**Algorithm2** Training GNN

**Input**: Fault to be detected *X*, Adjacency matrix *A*, Learning rate *η*, Number of iterations *t*.
**Output**: Matrix *Z*

1. Initialize $W^{(0)}$, $b^{(0)}$, $W^{(1)}$, $b^{(1)}$.
2. **for** iteration=1:*t*
3. hidden= $X * A * W^{(0)} * - b^{(0)}$.
4. $Z$= hidden $* A * W^{(1)} * - b^{(1)}$.
5. $loss = \frac{1}{2}(x' - z)^2$.
6. Update *W* and *b* using batch gradient decent $\nabla w(loss)$.
7. **end**
8. **return** *Z*.

### C. Fault detection

Definition 5: *Fault factor*

We define the fault factor of the *i*th data record $FF_i$ as the measure of the probability of fault. $FF_i$ is defined by the average reconstruction error over all features:

$$FF_i = \frac{1}{2}(X'_i - Z_i)^2 \quad (5)$$

The *FF* is evaluated for all data records by using the trained GNN. After the *FF* of each object is calculated and sorted, then the top-*n* objects with the largest fault factor are output as fault objects.

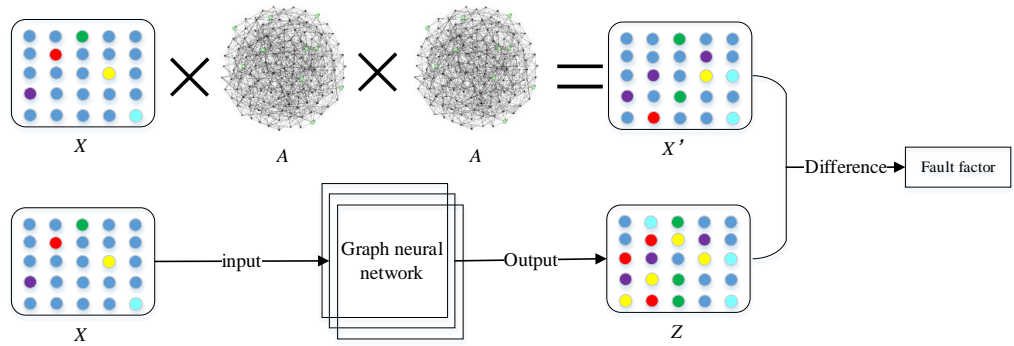

Figure. 4 Fault detection

**Algorithm3** Estimate the fault objects using the GNN

**Input**: Fault to be detected *X*, Adjacency matrix *A*, Matrix *Z*, *n* (the number of fault objects).
**Output**: The set of fault objects *F*.

1. Use equation (5) to calculate the *FF* of the object.
2. [value, index] = Sort (*FF*, 'descend').
3. *F*=index(1:*n*, :).
4. return *F*.

By jointly learning the feature of the object and its neighbors and then reconstructing them in the output layer of GNN, fault objects that are mixed in the normal object region or around the dense region will have larger *FF* and thus can be detected by GNN.

## III. EXPERIMENTAL INVESTIGATIONS

In order to verify the effectiveness of GNN method in bearing fault detection, we compare the GNN with several state-of-arts algorithms in Case Western Reserve University bearing data for experiments. The source code of our model is implemented in MATLAB 2019A. The hardware environment for the experiments is an Intel(R) Core(TM) i7-7700 3.60 GHz CPU with 16 GB of RAM. The operating system environment is Microsoft Windows 10 Professional.

### A. The Summary of Datasets and Compared Algorithms

Case Western Reserve University (CWRU) openly provides access to the collection of bearing fault detection datasets with ground truths. The proposed algorithm is validated using the CWRU dataset to illustrate the effectiveness of the proposed method.

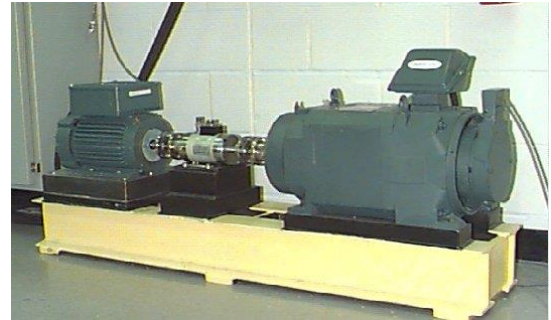

Figure. 5 Bearing test rig

#### 1) Data description

In the experimental investigations, the data are collected from the drive-end bearing. The bearing details are shown in Table 1.

Table 1. Parameter of the bearing

| Bearing type | Bearing location | Sample frequency(Hz) | Motor speed(rpm) | Motor load(HP) |
|---|---|---|---|---|
| SKF 6250 | Drive-end | 12000Hz | 1797 | 0 |

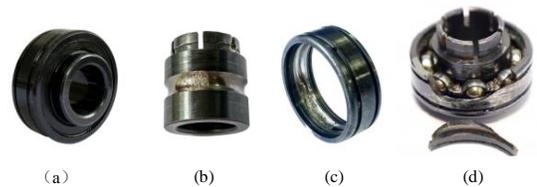

（a）　　　(b)　　　(c)　　　(d)

Figure. 6 Normal and fault state of bearings (a) normal (b) inner race fault (c) outer race fault (d) ball fault

Experimental data were divided into three groups of four states each: (i) 0.1778 mm inner race fault, 0.3556 mm inner race fault, 0.5334 mm inner race fault and normal base; (ii) 0.1778 mm ball fault, 0.3556 mm ball fault, 0.5334 mm ball fault and normal base; (iii) 0.1778 mm outer race fault, 0.3556 mm outer race fault, 0.5334 mm outer race fault and normal base.

In the present, the lengths of the normal samples are 240,000 data points, the lengths of the nine measurement fault samples are all 120,000 data points. Each sample is continuously divided into sub-samples with 300 data points. Twenty of each fault type are randomly selected according to different levels of fault diameters and mixed with normal samples to construct 3 groups of data sets.

Table 2. Summary of datasets

| Group 1 | | | Group 2 | | | Group 3 | | |
|---|---|---|---|---|---|---|---|---|
| Fault type | Fault diameter(mm) | Sample number | Fault type | Fault diameter(mm) | Sample number | Fault type | Fault diameter(mm) | Sample number |
| normal | 0 | 800 | normal | 0 | 800 | normal | 0 | 800 |
| inner race fault | 0.1778 | 20 | Ball fault | 0.1778 | 20 | Outer race fault | 0.1778 | 20 |
|  | 0.3556 | 20 |  | 0.3556 | 20 |  | 0.3556 | 20 |
|  | 0.5334 | 20 |  | 0.5334 | 20 |  | 0.5334 | 20 |
| Total |  | 860 | Total |  | 860 | Total |  | 860 |

The indicator set (also known as the feature set) is obtained by calculating 23 indicators in the time and frequency domains for the samples in the 3 group datasets, which are later used as input to the GNN model for fault detection to obtain the final detection results.

Table 3. Indexes and the calculation formulas

| Indexes | Formulas |
|---|---|
| 1.Standard deviation | $I_1 = \sqrt{\sum_{n=1}^{N}(x(n)-\bar{x})^2 / N}$ |
| 2.Peak | $I_2 = \max|x(n)|$ |
| 3.Skewness | $I_3 = \sum_{n=1}^{N}(x(n)-\bar{x})^3 / (N-1)I_1^3$ |
| 4.Kurtosis | $I_4 = \sum_{n=1}^{N}(x(n)-\bar{x})^4 / (N-1)I_1^4$ |
| 5.Root mean square | $I_5 = \sqrt{\sum_{n=1}^{N} x(n)^2 / N}$ |
| 6.Crest factor | $I_6 = I_2 / \sqrt{\sum_{n=1}^{N} x(n)^2 / N}$ |
| 7.Square | $I_7 = I_2 / (\sum_{n=1}^{N}\sqrt{x(n)}/N)^2$ |
| 8.Shape factor | $I_8 = \sqrt{N\sum_{n=1}^{N}x(n)^2} / \sum_{n=1}^{N}|x(n)|$ |
| 9.Impulse factor | $I_9 = \max|x(n)|/(\sum_{n=1}^{N}|x(n)|/N)$ |
| 10.WPD energy | $I_{10} = \sum_{n=1}^{N}|x_i(n)|^2 / \sum_{i=0}^{2^{j-1}}\sum_{n=1}^{N}|x_i(n)|^2$ |
| 11.EEMD energy | $I_{11} = \sum_{n=1}^{N}|IMF_i(n)|^2 / \sum_{i=1}^{NI}\sum_{n=1}^{N}|IMF_i(n)|^2$ |

$x(n)$ and $\bar{x}$ denote the data sequence and mean of the data sequence, respectively; $N$ is the number of the data points. $x_i(n)$ is the decomposition coefficient sequence of the *i*th ($i=0,1,…,2^j-1$, $j$ is the WPD decomposition level) frequency band using WPD; $IMF_i(n)$ is the *i*th data sequence after EEMD, and $NI$ is the decomposition level using EEMD.

Based on Table 3, the index is calculated for each sample. Four steps are required.

1): Nine time domain indexes are calculated as follows:

$$I=[I_1,I_2,I_3,I_4,I_5,I_6,I_7,I_8,I_9] \quad (6)$$

2): $E_{WPD}$ is obtained by calculating WPD energy (parameters $j = 3$ and wavelet Db20).

$$W_{WPD}=[E_{WPD}^1,E_{WPD}^2,E_{WPD}^3,E_{WPD}^4,E_{WPD}^5,E_{WPD}^6,E_{WPD}^7,E_{WPD}^8] \quad (7)$$

3): EEMD energy is calculated to obtain a dataset as follows:

$$W_{EEMD}=[E_{EEMD}^1,E_{EEMD}^2,E_{EEMD}^3,E_{EEMD}^4,E_{EEMD}^5,E_{EEMD}^6] \quad (8)$$

4): $I$, $W_{WPD}$, $W_{EEMD}$ are combined into a dataset as follows:

$$X=[I,W_{WPD},W_{EEMD}] \quad (9)$$

*2) Compared Algorithms*

We selected three different types of state-of-the-art outlier detection algorithms for comparison experiments with the proposed GNN. These algorithms are common types in the outlier detection field, and they are used as comparison algorithms in most of the related literature.

Table 4. Comparison algorithm statistics

| Type of algorithm | Acronym of algorithm |
|---|---|
| Neuron network-based | AE |
| Local outlier factor-based | LOF |
| Connective-based | COF |

Due to the parameter settings for each type of algorithm are different. Therefore, Table 5 is used to describe in detail the parameter settings of each algorithm in the experiments.

Table 5. Parameter Setting

| Algorithms | k(Number of nearest neighbors) | Learning rate | Number of iterations | Number of layers |
|---|---|---|---|---|
| GNN | 2~10 | 0.0001~0.002 | 10~100 | 3 |
| AE | \ | 0.0001~0.002 | 10~100 | 3 |
| LOF | 2~100 | \ | \ | \ |
| COF | 2~100 | \ | \ | \ |

*B. Evaluation Techniques*

In real-world applications, ground truth faults are generally rare. The Receiver Operating Characteristic (ROC) curve, which captures the trade-off between the sensitivity and specificity, has been adopted by a large proportion of literatures in this area. An algorithm with a large AUC value is preferred. We choose the AUC, ACC (Accuracy), DR (Detection Rate), FAR (False Alarm Rate) as the algorithm performance evaluation metric. Higher AUC, ACC, and DR values and lower FAR, execution time indicate better performance of the algorithm.

Let *TP* be the number of fault objects correctly marked as faults by the algorithm; *TN* be the number of normal objects correctly marked as normal by the algorithm; *FP* be the number of normal objects incorrectly marked as faults by the algorithm; *FN* be the number of faults that the algorithm incorrectly marks as normal objects. The calculation method of each evaluation techniques are shown in Algorithm 4.

**Algorithm 4** Evaluation Techniques

1. Let $n_o$ be the number of true faults.
2. Let $n_n$ be the number of true normal objects.
3. Let *S* be the sum of ranking of the actual faults,

$S=\sum_{i=1}^{n_o} r_i$ , where $r_i$ is the rank of the *i*th outlier in the ranked list.

4. $AUC = \dfrac{S - (n_o^2 + n_o)/2}{n_o n_n}$

5. $ACC = \dfrac{TP + TN}{TP + TN + FP + FN}$

6. $DR = \dfrac{TP}{TP + FN}$

7. $FAR = \dfrac{FP}{TN + FP}$

### C. Experiment results

We judged the *n* objects(The size of *n* is equal to the number of true faults in the dataset) with the highest fault factor determined by the GNN as fault objects, and compared them with the labels, then plotted the GNN detection results in Fig 5.

We use the blue hollow circles represent normal objects, the green solid diamonds represent the true fault objects in each group, and the red hollow diamonds represent the fault objects determined by GNN. If the green solid diamond is in the red diamond, it means that the GNN is judged correctly. If the blue hollow circles are in the red diamond, it means that the GNN is judged incorrectly.

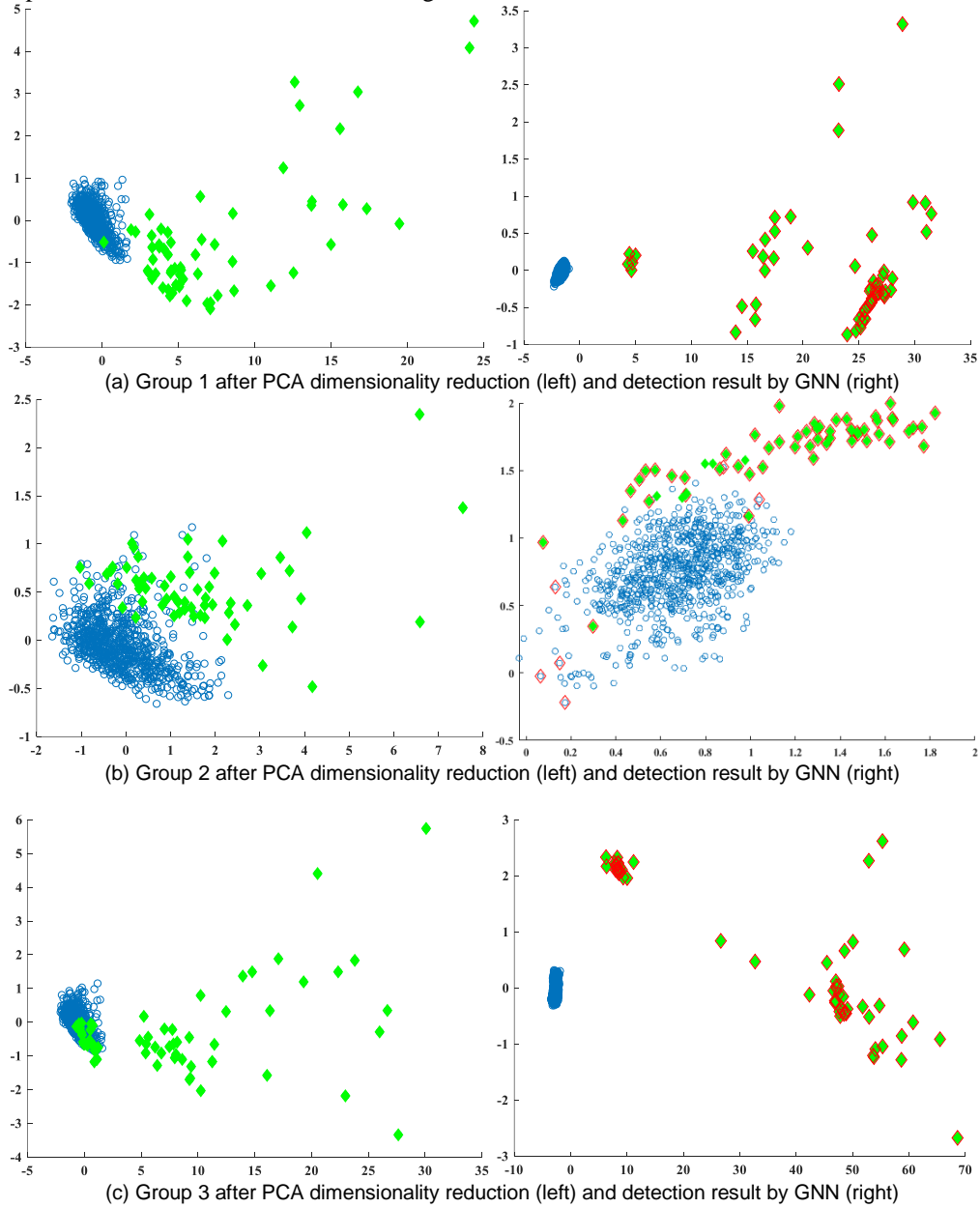

(a) Group 1 after PCA dimensionality reduction (left) and detection result by GNN (right)

(b) Group 2 after PCA dimensionality reduction (left) and detection result by GNN (right)

(c) Group 3 after PCA dimensionality reduction (left) and detection result by GNN (right)

Figure. 7 The original data distribution after PCA and detection result of GNN

On the left side of Fig 7, some of the fault objects are mixed with the normal objects. Usually, the reason for this phenomenon is that the fault characteristics are not obvious and are in the early stage of the fault, which is difficult to detect by most algorithms. On the right side of Fig 7, the distribution of the data is different from the left side, fault objects deviate more from normal objects. The main reason for this difference is GNN aggregates objects neighbor feature values at each layer of the network, making the distribution pattern of the data changed, thus enabling fault objects originally hidden in the normal objects area to be detected.

The GNN method is compared with AE, LOF, COF, for each algorithm on each group, we adjust the parameters to run 10 experiments and select the best detection result as the final performance of the algorithm.

Table 6. Experimental results. Values marked in bold are ranked by the top 2 in the dataset.

| Datasets | AUC | | | |
|---|---|---|---|---|
| | GNN | AE | LOF | COF |
| Group 1 | **1** | 0.93 | 0.71 | **0.94** |

| Datasets | | | | |
|---|---|---|---|---|
| Group 2 | **0.99** | **0.87** | 0.73 | 0.83 |
| Group 3 | **1** | 0.91 | **0.94** | 0.60 |
| (a) AUC performance | | | | |
| Datasets | ACC (%) | | | |
| | GNN | AE | LOF | COF |
| Group 1 | **100** | 97.20 | 93.72 | **97.67** |
| Group 2 | **98.61** | 92.79 | 91.62 | **92.79** |
| Group 3 | **100** | 95.34 | 96.27 | 91.62 |
| (b) ACC performance | | | | |
| Datasets | DR (%) | | | |
| | GNN | AE | LOF | COF |
| Group 1 | **100** | 80 | 55 | **83.33** |
| Group 2 | **90** | **48.33** | 40 | 48.33 |
| Group 3 | **100** | 66.67 | 73.33 | 40 |
| (c) Detection rate performance (%) | | | | |
| Datasets | FAR (%) | | | |
| | GNN | AE | LOF | COF |
| Group 1 | **0** | 1.50 | 3.37 | **1.25** |
| Group 2 | **0.75** | 3.87 | 4.50 | **3.87** |
| Group 3 | **0** | 2.50 | **2.00** | 4.50 |
| (d) False alarm rate (%) | | | | |
| Datasets | Time(seconds) | | | |
| | GNN | AE | LOF | COF |
| Group 1 | **0.44** | 0.31 | 2.03 | 1.46 |
| Group 2 | **0.45** | 0.35 | 2.11 | 1.69 |
| Group 3 | **0.46** | 0.34 | 2.12 | 1.59 |
| (e) Actual execution time | | | | |

The GNN method achieves the best detection performance among the four evaluation metrics of AUC, ACC, DR, FAR, but its execution time is slightly higher than that of AE. This is because the detection process of GNN has graphs involved in the computation, which improves the computational overhead. It is demonstrated that GNN is effective in detecting faults in the dataset, including those early fault signals that are difficult to be detected by other algorithms.

Each iteration of the GNN network is fitting not only the feature information of the objects themselves, but also the feature information of the objects and their neighbors. At the same time, since normal objects are more similar to each other, this makes the reconstruct error of normal objects tend to be similar in each iteration of the network. The fault objects, on the other hand, as they deviate from the majority of normal objects, their reconstruction error gradually increases compared to normal objects as the network iterations. Therefore, when the GNN network stops iterating, it will be easy to distinguish normal objects from fault objects by comparing the reconstruction error of each object.

### D. Research the Influence of Parameters on Detection Performance of GNN

We conducted 20 experiments on each dataset to investigate the effects of the number of neighbor's *k*, the number of layers on the performance of the GNN. The experimental results are shown in below.

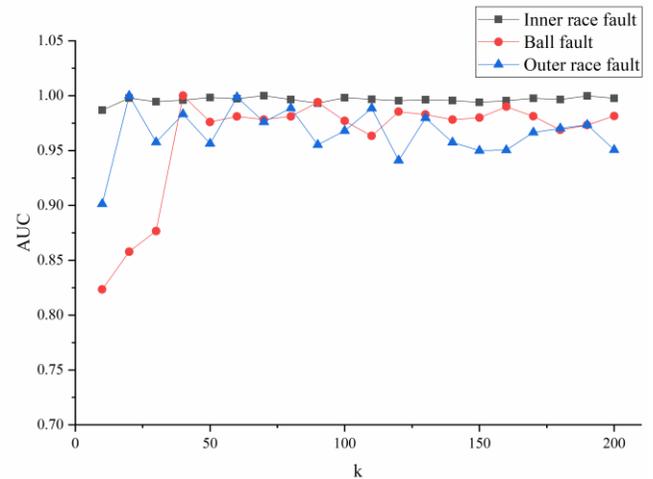

Figure. 8 The influence of the number of nearest neighbors *k* on GNN

From figure 8, we can see that when the nearest neighbor's *k* gradually increases, the AUC of GNN also gradually increases and then tend to stable. The main reason for this phenomenon is when the value of *k* is small, the hidden layer of the network aggregates less information about the neighborhood features of the objects and cannot distinguish significantly between normal and fault objects.

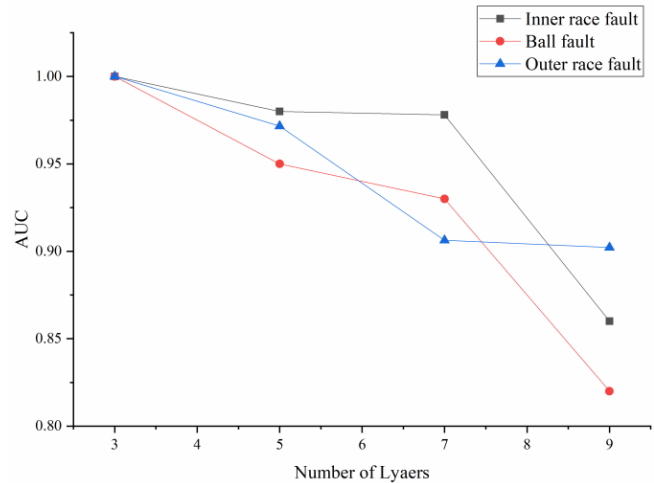

Figure. 9 The influence of the number of layers on GNN

As the depth of the network deepens, the feature values between objects become more and more similar, their reconstruction errors in the output layer of the network to become more similar. That making it difficult to distinguish normal objects from fault objects and decreased the performance of the GNN algorithm.

## IV. CONCLUSION

In this paper, a novel fault detection scheme of graph neural network-based is proposed to solve the early fault detection problems. Due to the small amplitude and intensity of the characteristic signal of the early fault, its characteristics are extremely inconspicuous, random and hidden, easily masked by system interference and noise, and difficult to detect by traditional methods. Aim to this problem, we design a novel graph neural network structure specifically for unsupervised fault detection. This method first converts the vibration signals into graph, so that the vibration signals, which are originally independent of each other, are correlated with each other; then inputs the dataset together with its corresponding graph into the GNN for training; and finally determines the top-*n* objects that are difficult to reconstruct in the output layer of the GNN as fault objects. By comparing GNN with several state of art algorithms in detail, we find that the proposed method can successfully detect fault objects that are mixed in the normal object region. In future work, we attempt to introduce ensemble learning into GNN to achieve more satisfactory and stable results.